\documentclass[submission,copyright,creativecommons,sharealike,noncommercial]{eptcs}
\providecommand{\event}{Workshop on Semantic Spaces at the\\ Intersection of NLP, Physics and Cognitive Science}
\pdfoutput=1

\usepackage{mathtools} 
\usepackage{amssymb} 
\usepackage{amsthm} 

\usepackage{relsize} 
\usepackage{microtype} 

\usepackage[nocompress]{cite} 

\usepackage{graphicx} 
\usepackage[usenames,dvipsnames]{xcolor} 
\usepackage{tikz} 
\usepackage{tikzfig} 

		\numberwithin{theorem_c}{section} 
		\numberwithin{equation}{section} 

		\theoremstyle{plain}

		\theoremstyle{definition}

	\newcommand{\naturals}{\mathbb{N}} 
	\newcommand{\integers}{\mathbb{Z}} 
	\newcommand{\reals}{\mathbb{R}} 
	\newcommand{\complexs}{\mathbb{C}} 
	
	\newcommand{\suchthat}[2]{\left\{#1 \: \middle\vert \: #2\right\}} 

		\newcommand{\ket}[1]{\vert #1 \rangle} 
		\newcommand{\braket}[2]{\langle #1 \vert #2 \rangle} 
		
		\newcommand{\SpaceH}{\mathcal{H}}

		\newcommand{\id}[1]{id_{#1}} 

		\newcommand{\RModCategory}[1]{#1\operatorname{-Mod}} 

	\newcommand{\Zbwcolour}{white}

	\tikzset{->-/.style={decoration={markings,mark=at position #1 with {\arrow{>}}},postaction={decorate}}}
	\tikzset{-<-/.style={decoration={markings,mark=at position #1 with {\arrow{<}}},postaction={decorate}}}

\tikzstyle{env}=[copoint,regular polygon rotate=0,minimum width=0.2cm, fill=black]

\tikzstyle{probs}=[shape=semicircle,fill=white,draw=black,shape border rotate=180,minimum width=1.2cm]

\tikzstyle{every picture}=[baseline=-0.25em,scale=0.5]
\tikzstyle{dotpic}=[] 
\tikzstyle{diredges}=[every to/.style={diredge}]
\tikzstyle{math matrix}=[matrix of math nodes,left delimiter=(,right delimiter=),inner sep=2pt,column sep=1em,row sep=0.5em,nodes={inner sep=0pt},text height=1.5ex, text depth=0.25ex]

\tikzstyle{inline text}=[text height=1.5ex, text depth=0.25ex,yshift=0.5mm]
\tikzstyle{label}=[font=\footnotesize,text height=1.5ex, text depth=0.25ex,yshift=0.5mm]
\tikzstyle{left label}=[label,anchor=east,xshift=1.5mm]
\tikzstyle{right label}=[label,anchor=west,xshift=-1.5mm]

\tikzstyle{braceedge}=[decorate,decoration={brace,amplitude=2mm,raise=-1mm}]
\tikzstyle{small braceedge}=[decorate,decoration={brace,amplitude=1mm,raise=-1mm}]

\tikzstyle{doubled}=[line width=1.6pt] 
\tikzstyle{boldedge}=[doubled,shorten <=-0.17mm,shorten >=-0.17mm]
\tikzstyle{boldedgegray}=[doubled,gray,shorten <=-0.17mm,shorten >=-0.17mm]

\tikzstyle{semidoubled}=[line width=1.4pt] 
\tikzstyle{semiboldedgegray}=[semidoubled,gray,shorten <=-0.17mm,shorten >=-0.17mm]

\tikzstyle{boldedgedashed}=[very thick,dashed,shorten <=-0.17mm,shorten >=-0.17mm]
\tikzstyle{vboldedgedashed}=[doubled,dashed,shorten <=-0.17mm,shorten >=-0.17mm]
\tikzstyle{left hook arrow}=[left hook-latex]
\tikzstyle{right hook arrow}=[right hook-latex]
\tikzstyle{sembracket}=[line width=0.5pt,shorten <=-0.07mm,shorten >=-0.07mm]

\tikzstyle{causal edge}=[->,thick,gray]
\tikzstyle{causal nondir}=[thick,gray]
\tikzstyle{timeline}=[thick,gray, dashed]

\tikzstyle{cedge}=[<->,thick,gray!70!white]

\tikzstyle{empty diagram}=[draw=gray!40!white,dashed,shape=rectangle,minimum width=1cm,minimum height=1cm]
\tikzstyle{empty diagram small}=[draw=gray!50!white,dashed,shape=rectangle,minimum width=0.6cm,minimum height=0.5cm]

\tikzstyle{dot}=[inner sep=0mm,minimum width=3mm,minimum height=3mm,draw,shape=circle,text depth=-0.1mm]
\tikzstyle{ddot}=[inner sep=0mm, doubled, minimum width=3.5mm,minimum height=3.5mm,draw,shape=circle]

\tikzstyle{black dot}=[dot,fill=black]
\tikzstyle{white dot}=[dot,fill=white,,text depth=-0.2mm]
\tikzstyle{green dot}=[white dot] 
\tikzstyle{gray dot}=[dot,fill=gray!40!white,,text depth=-0.2mm]
\tikzstyle{red dot}=[gray dot] 

\tikzstyle{black ddot}=[ddot,fill=black]
\tikzstyle{white ddot}=[ddot,fill=white]
\tikzstyle{gray ddot}=[ddot,fill=gray!40!white]

\tikzstyle{gray edge}=[gray!40!white]

\tikzstyle{small dot}=[inner sep=0.5mm,minimum width=0pt,minimum height=0pt,draw,shape=circle]

\tikzstyle{small black dot}=[small dot,fill=black]
\tikzstyle{small white dot}=[small dot,fill=white]
\tikzstyle{small gray dot}=[small dot,fill=gray!40!white]

\tikzstyle{causal dot}=[inner sep=0.4mm,minimum width=0pt,minimum height=0pt,draw=white,shape=circle,fill=gray!40!white]

\tikzstyle{phase dimensions}=[minimum size=5mm,font=\footnotesize,rectangle,rounded corners=2.5mm,inner sep=0.2mm,outer sep=-2mm,text height=1ex, text depth=0.25ex, yshift=0.5mm]
\tikzstyle{dphase dimensions}=[phase dimensions]

\tikzstyle{phase dot}=[dot,phase dimensions]

\tikzstyle{white phase dot}=[dot,fill=white,phase dimensions]
\tikzstyle{white phase ddot}=[ddot,fill=white,dphase dimensions]

\tikzstyle{white rect ddot}=[draw=black,fill=white,doubled,minimum size=5mm,font=\footnotesize,rectangle,rounded corners=2.5mm,inner sep=0.2mm]
\tikzstyle{gray rect ddot}=[draw=black,fill=gray!40!white,doubled,minimum size=6mm,font=\footnotesize,rectangle,rounded corners=3mm]

\tikzstyle{gray phase dot}=[dot,fill=gray!40!white,phase dimensions]
\tikzstyle{gray phase ddot}=[ddot,fill=gray!40!white,dphase dimensions]
\tikzstyle{grey phase dot}=[gray phase dot]
\tikzstyle{grey phase ddot}=[gray phase ddot]

\tikzstyle{cnot}=[fill=white,shape=circle,inner sep=-1.4pt]
\tikzstyle{hadamard}=[square box,inner sep=0 pt,font=\footnotesize,minimum height=4mm,minimum width=4mm]
\tikzstyle{dhadamard}=[hadamard,doubled]
\tikzstyle{antipode}=[white dot,inner sep=0.3mm,font=\footnotesize]

\tikzstyle{scalar}=[diamond,draw,inner sep=0.5pt,font=\small]
\tikzstyle{dscalar}=[diamond,doubled, draw,inner sep=0.5pt,font=\small]

\tikzstyle{small box}=[rectangle,inline text,fill=white,draw,minimum height=5mm,yshift=-0.5mm,minimum width=5mm,font=\small]
\tikzstyle{small gray box}=[small box,fill=gray!30]
\tikzstyle{medium box}=[rectangle,inline text,fill=white,draw,minimum height=5mm,yshift=-0.5mm,minimum width=10mm,font=\small]
\tikzstyle{square box}=[small box] 
\tikzstyle{medium gray box}=[small box,fill=gray!30]
\tikzstyle{semilarge box}=[rectangle,inline text,fill=white,draw,minimum height=5mm,yshift=-0.5mm,minimum width=12.5mm,font=\small]
\tikzstyle{large box}=[rectangle,inline text,fill=white,draw,minimum height=5mm,yshift=-0.5mm,minimum width=15mm,font=\small]
\tikzstyle{large gray box}=[small box,fill=gray!30]

\tikzstyle{gray square point}=[small box,fill=gray!50]

\tikzstyle{dphase box white}=[dbox]
\tikzstyle{dphase box gray}=[dbox,fill=gray!50!white]

\tikzstyle{point}=[regular polygon,regular polygon sides=3,draw,scale=0.75,inner sep=-0.5pt,minimum width=9mm,fill=white,regular polygon rotate=180]
\tikzstyle{copoint}=[regular polygon,regular polygon sides=3,draw,scale=0.75,inner sep=-0.5pt,minimum width=9mm,fill=white]
\tikzstyle{dpoint}=[point,doubled]
\tikzstyle{dcopoint}=[copoint,doubled]

\tikzstyle{wide copoint}=[fill=white,draw,shape=isosceles triangle,shape border rotate=90,isosceles triangle stretches=true,inner sep=0pt,minimum width=1.5cm,minimum height=6.12mm]
\tikzstyle{wide point}=[fill=white,draw,shape=isosceles triangle,shape border rotate=-90,isosceles triangle stretches=true,inner sep=0pt,minimum width=1.5cm,minimum height=6.12mm,yshift=-0.0mm]
\tikzstyle{wide point plus}=[fill=white,draw,shape=isosceles triangle,shape border rotate=-90,isosceles triangle stretches=true,inner sep=0pt,minimum width=1.74cm,minimum height=7mm,yshift=-0.0mm]

\tikzstyle{wide dpoint}=[fill=white,doubled,draw,shape=isosceles triangle,shape border rotate=-90,isosceles triangle stretches=true,inner sep=0pt,minimum width=1.5cm,minimum height=6.12mm,yshift=-0.0mm]

\tikzstyle{tinypoint}=[regular polygon,regular polygon sides=3,draw,scale=0.55,inner sep=-0.15pt,minimum width=6mm,fill=white,regular polygon rotate=180] 

\tikzstyle{white point}=[point]
\tikzstyle{white dpoint}=[dpoint]
\tikzstyle{green point}=[white point] 
\tikzstyle{white copoint}=[copoint]
\tikzstyle{gray point}=[point,fill=gray!40!white]
\tikzstyle{gray dpoint}=[gray point,doubled]
\tikzstyle{red point}=[gray point] 
\tikzstyle{gray copoint}=[copoint,fill=gray!40!white]
\tikzstyle{gray dcopoint}=[gray copoint,doubled]

\tikzstyle{black point}=[point,fill=black]
\tikzstyle{black copoint}=[copoint,fill=black]

\tikzstyle{tiny gray point}=[tinypoint,fill=gray!40!white]

\tikzstyle{diredge}=[->]
\tikzstyle{rdiredge}=[<-]
\tikzstyle{thickdiredge}=[->, very thick]
\tikzstyle{pointer edge}=[->,very thick,gray]
\tikzstyle{pointer edge part}=[very thick,gray]
\tikzstyle{dashed edge}=[dashed]
\tikzstyle{thick dashed edge}=[very thick,dashed]
\tikzstyle{thick gray dashed edge}=[thick dashed edge,gray!40]
\tikzstyle{thick map edge}=[very thick,|->]

\makeatletter
\newcommand{\boxshape}[3]{%
\pgfdeclareshape{#1}{
\inheritsavedanchors[from=rectangle] 
\inheritanchorborder[from=rectangle]
\inheritanchor[from=rectangle]{center}
\inheritanchor[from=rectangle]{north}
\inheritanchor[from=rectangle]{south}
\inheritanchor[from=rectangle]{west}
\inheritanchor[from=rectangle]{east}
\backgroundpath{
\southwest \pgf@xa=\pgf@x \pgf@ya=\pgf@y
\northeast \pgf@xb=\pgf@x \pgf@yb=\pgf@y

\@tempdima=#2
\@tempdimb=#3

\pgfpathmoveto{\pgfpoint{\pgf@xa - 5pt + \@tempdima}{\pgf@ya}}
\pgfpathlineto{\pgfpoint{\pgf@xa - 5pt - \@tempdima}{\pgf@yb}}
\pgfpathlineto{\pgfpoint{\pgf@xb + 5pt + \@tempdimb}{\pgf@yb}}
\pgfpathlineto{\pgfpoint{\pgf@xb + 5pt - \@tempdimb}{\pgf@ya}}
\pgfpathlineto{\pgfpoint{\pgf@xa - 5pt + \@tempdima}{\pgf@ya}}
\pgfpathclose
}
}}

\boxshape{NEbox}{0pt}{5pt}
\boxshape{SEbox}{0pt}{-5pt}
\boxshape{NWbox}{5pt}{0pt}
\boxshape{SWbox}{-5pt}{0pt}
\boxshape{EBox}{-3pt}{3pt}
\boxshape{WBox}{3pt}{-3pt}
\makeatother

\tikzstyle{cloud}=[shape=cloud,draw,minimum width=1.5cm,minimum height=1.5cm]

\tikzstyle{map}=[draw,shape=NEbox,inner sep=2pt,minimum height=6mm,fill=white]
\tikzstyle{dashedmap}=[draw,dashed,shape=NEbox,inner sep=2pt,minimum height=6mm,fill=white]
\tikzstyle{mapdag}=[draw,shape=SEbox,inner sep=2pt,minimum height=6mm,fill=white]
\tikzstyle{mapadj}=[draw,shape=SEbox,inner sep=2pt,minimum height=6mm,fill=white]
\tikzstyle{maptrans}=[draw,shape=SWbox,inner sep=2pt,minimum height=6mm,fill=white]
\tikzstyle{mapconj}=[draw,shape=NWbox,inner sep=2pt,minimum height=6mm,fill=white]

\tikzstyle{medium map}=[draw,shape=NEbox,inner sep=2pt,minimum height=6mm,fill=white,minimum width=7mm]
\tikzstyle{medium map dag}=[draw,shape=SEbox,inner sep=2pt,minimum height=6mm,fill=white,minimum width=7mm]
\tikzstyle{medium map adj}=[draw,shape=SEbox,inner sep=2pt,minimum height=6mm,fill=white,minimum width=7mm]
\tikzstyle{medium map trans}=[draw,shape=SWbox,inner sep=2pt,minimum height=6mm,fill=white,minimum width=7mm]
\tikzstyle{medium map conj}=[draw,shape=NWbox,inner sep=2pt,minimum height=6mm,fill=white,minimum width=7mm]
\tikzstyle{semilarge map}=[draw,shape=NEbox,inner sep=2pt,minimum height=6mm,fill=white,minimum width=9.5mm]
\tikzstyle{semilarge map trans}=[draw,shape=SWbox,inner sep=2pt,minimum height=6mm,fill=white,minimum width=9.5mm]
\tikzstyle{semilarge map adj}=[draw,shape=SEbox,inner sep=2pt,minimum height=6mm,fill=white,minimum width=9.5mm]
\tikzstyle{semilarge map dag}=[draw,shape=SEbox,inner sep=2pt,minimum height=6mm,fill=white,minimum width=9.5mm]
\tikzstyle{semilarge map conj}=[draw,shape=NWbox,inner sep=2pt,minimum height=6mm,fill=white,minimum width=9.5mm]
\tikzstyle{large map}=[draw,shape=NEbox,inner sep=2pt,minimum height=6mm,fill=white,minimum width=12mm]
\tikzstyle{very large map}=[draw,shape=NEbox,inner sep=2pt,minimum height=6mm,fill=white,minimum width=17mm]

\tikzstyle{medium dmap}=[draw,doubled,shape=NEbox,inner sep=2pt,minimum height=6mm,fill=white,minimum width=7mm]
\tikzstyle{medium dmap dag}=[draw,doubled,shape=SEbox,inner sep=2pt,minimum height=6mm,fill=white,minimum width=7mm]
\tikzstyle{medium dmap adj}=[draw,doubled,shape=SEbox,inner sep=2pt,minimum height=6mm,fill=white,minimum width=7mm]
\tikzstyle{medium dmap trans}=[draw,doubled,shape=SWbox,inner sep=2pt,minimum height=6mm,fill=white,minimum width=7mm]
\tikzstyle{medium dmap conj}=[draw,doubled,shape=NWbox,inner sep=2pt,minimum height=6mm,fill=white,minimum width=7mm]
\tikzstyle{semilarge dmap}=[draw,doubled,shape=NEbox,inner sep=2pt,minimum height=6mm,fill=white,minimum width=9.5mm]
\tikzstyle{semilarge dmap trans}=[draw,doubled,shape=SWbox,inner sep=2pt,minimum height=6mm,fill=white,minimum width=9.5mm]
\tikzstyle{semilarge dmap adj}=[draw,doubled,shape=SEbox,inner sep=2pt,minimum height=6mm,fill=white,minimum width=9.5mm]
\tikzstyle{semilarge dmap dag}=[draw,doubled,shape=SEbox,inner sep=2pt,minimum height=6mm,fill=white,minimum width=9.5mm]
\tikzstyle{semilarge dmap conj}=[draw,doubled,shape=NWbox,inner sep=2pt,minimum height=6mm,fill=white,minimum width=9.5mm]
\tikzstyle{large dmap}=[draw,doubled,shape=NEbox,inner sep=2pt,minimum height=6mm,fill=white,minimum width=12mm]
\tikzstyle{large dmap conj}=[draw,doubled,shape=NWbox,inner sep=2pt,minimum height=6mm,fill=white,minimum width=12mm]
\tikzstyle{large dmap trans}=[draw,doubled,shape=SWbox,inner sep=2pt,minimum height=6mm,fill=white,minimum width=12mm]
\tikzstyle{very large dmap}=[draw,doubled,shape=NEbox,inner sep=2pt,minimum height=6mm,fill=white,minimum width=19.5mm]

\tikzstyle{muxbox}=[draw,shape=rectangle,minimum height=3mm,minimum width=3mm,fill=white]
\tikzstyle{dmuxbox}=[muxbox,doubled]

\tikzstyle{box}=[draw,shape=rectangle,inner sep=2pt,minimum height=6mm,minimum width=6mm,fill=white]
\tikzstyle{dbox}=[draw,doubled,shape=rectangle,inner sep=2pt,minimum height=6mm,minimum width=6mm,fill=white]
\tikzstyle{dmap}=[draw,doubled,shape=NEbox,inner sep=2pt,minimum height=6mm,fill=white]
\tikzstyle{dmapdag}=[draw,doubled,shape=SEbox,inner sep=2pt,minimum height=6mm,fill=white]
\tikzstyle{dmapadj}=[draw,doubled,shape=SEbox,inner sep=2pt,minimum height=6mm,fill=white]
\tikzstyle{dmaptrans}=[draw,doubled,shape=SWbox,inner sep=2pt,minimum height=6mm,fill=white]
\tikzstyle{dmapconj}=[draw,doubled,shape=NWbox,inner sep=2pt,minimum height=6mm,fill=white]

\tikzstyle{ddmap}=[draw,doubled,dashed,shape=NEbox,inner sep=2pt,minimum height=6mm,fill=white]
\tikzstyle{ddmapdag}=[draw,doubled,dashed,shape=SEbox,inner sep=2pt,minimum height=6mm,fill=white]
\tikzstyle{ddmapadj}=[draw,doubled,dashed,shape=SEbox,inner sep=2pt,minimum height=6mm,fill=white]
\tikzstyle{ddmaptrans}=[draw,doubled,dashed,shape=SWbox,inner sep=2pt,minimum height=6mm,fill=white]
\tikzstyle{ddmapconj}=[draw,doubled,dashed,shape=NWbox,inner sep=2pt,minimum height=6mm,fill=white]

\boxshape{sNEbox}{0pt}{3pt}
\boxshape{sSEbox}{0pt}{-3pt}
\boxshape{sNWbox}{3pt}{0pt}
\boxshape{sSWbox}{-3pt}{0pt}
\tikzstyle{smap}=[draw,shape=sNEbox,fill=white]
\tikzstyle{smapdag}=[draw,shape=sSEbox,fill=white]
\tikzstyle{smapadj}=[draw,shape=sSEbox,fill=white]
\tikzstyle{smaptrans}=[draw,shape=sSWbox,fill=white]
\tikzstyle{smapconj}=[draw,shape=sNWbox,fill=white]

\tikzstyle{dsmap}=[draw,dashed,shape=sNEbox,fill=white]
\tikzstyle{dsmapdag}=[draw,dashed,shape=sSEbox,fill=white]
\tikzstyle{dsmaptrans}=[draw,dashed,shape=sSWbox,fill=white]
\tikzstyle{dsmapconj}=[draw,dashed,shape=sNWbox,fill=white]

\boxshape{mNEbox}{0pt}{10pt}
\boxshape{mSEbox}{0pt}{-10pt}
\boxshape{mNWbox}{10pt}{0pt}
\boxshape{mSWbox}{-10pt}{0pt}
\tikzstyle{mmap}=[draw,shape=mNEbox]
\tikzstyle{mmapdag}=[draw,shape=mSEbox]
\tikzstyle{mmaptrans}=[draw,shape=mSWbox]
\tikzstyle{mmapconj}=[draw,shape=mNWbox]

\tikzstyle{mmapgray}=[draw,fill=gray!40!white,shape=mNEbox]
\tikzstyle{smapgray}=[draw,fill=gray!40!white,shape=sNEbox]

\makeatletter
\pgfdeclareshape{cornerpoint}{
\inheritsavedanchors[from=rectangle] 
\inheritanchorborder[from=rectangle]
\inheritanchor[from=rectangle]{center}
\inheritanchor[from=rectangle]{north}
\inheritanchor[from=rectangle]{south}
\inheritanchor[from=rectangle]{west}
\inheritanchor[from=rectangle]{east}
\backgroundpath{
\southwest \pgf@xa=\pgf@x \pgf@ya=\pgf@y
\northeast \pgf@xb=\pgf@x \pgf@yb=\pgf@y

\pgfmathsetmacro{\pgf@shorten@left}{\pgfkeysvalueof{/tikz/shorten left}}
\pgfmathsetmacro{\pgf@shorten@right}{\pgfkeysvalueof{/tikz/shorten right}}

\pgfpathmoveto{\pgfpoint{0.5 * (\pgf@xa + \pgf@xb)}{\pgf@ya - 5pt}}
\pgfpathlineto{\pgfpoint{\pgf@xa - 8pt + \pgf@shorten@left}{\pgf@yb - 1.5 * \pgf@shorten@left}}
\pgfpathlineto{\pgfpoint{\pgf@xa - 8pt + \pgf@shorten@left}{\pgf@yb}}
\pgfpathlineto{\pgfpoint{\pgf@xb + 8pt - \pgf@shorten@right}{\pgf@yb}}
\pgfpathlineto{\pgfpoint{\pgf@xb + 8pt - \pgf@shorten@right}{\pgf@yb - 1.5 * \pgf@shorten@right}}
\pgfpathclose
}
}

\pgfdeclareshape{cornercopoint}{
\inheritsavedanchors[from=rectangle] 
\inheritanchorborder[from=rectangle]
\inheritanchor[from=rectangle]{center}
\inheritanchor[from=rectangle]{north}
\inheritanchor[from=rectangle]{south}
\inheritanchor[from=rectangle]{west}
\inheritanchor[from=rectangle]{east}
\backgroundpath{
\southwest \pgf@xa=\pgf@x \pgf@ya=\pgf@y
\northeast \pgf@xb=\pgf@x \pgf@yb=\pgf@y

\pgfmathsetmacro{\pgf@shorten@left}{\pgfkeysvalueof{/tikz/shorten left}}
\pgfmathsetmacro{\pgf@shorten@right}{\pgfkeysvalueof{/tikz/shorten right}}

\pgfpathmoveto{\pgfpoint{0.5 * (\pgf@xa + \pgf@xb)}{\pgf@yb + 5pt}}
\pgfpathlineto{\pgfpoint{\pgf@xa - 8pt + \pgf@shorten@left}{\pgf@ya + 1.5 * \pgf@shorten@left}}
\pgfpathlineto{\pgfpoint{\pgf@xa - 8pt + \pgf@shorten@left}{\pgf@ya}}
\pgfpathlineto{\pgfpoint{\pgf@xb + 8pt - \pgf@shorten@right}{\pgf@ya}}
\pgfpathlineto{\pgfpoint{\pgf@xb + 8pt - \pgf@shorten@right}{\pgf@ya + 1.5 * \pgf@shorten@right}}
\pgfpathclose
}
}

\makeatother

\pgfkeyssetvalue{/tikz/shorten left}{0pt}
\pgfkeyssetvalue{/tikz/shorten right}{0pt}

\tikzstyle{kpoint common}=[draw,fill=white,inner sep=1pt,minimum height=3mm]
\tikzstyle{kpoint}=[shape=cornerpoint,shorten left=5pt,kpoint common]
\tikzstyle{kpoint adjoint}=[shape=cornercopoint,shorten left=5pt,kpoint common]
\tikzstyle{kpoint conjugate}=[shape=cornerpoint,shorten right=5pt,kpoint common]
\tikzstyle{kpoint transpose}=[shape=cornercopoint,shorten right=5pt,kpoint common]
\tikzstyle{kpoint symm}=[shape=cornerpoint,shorten left=5pt,shorten right=5pt,kpoint common]

\tikzstyle{black kpoint}=[shape=cornerpoint,shorten left=5pt,kpoint common,fill=black]
\tikzstyle{black kpoint adjoint}=[shape=cornercopoint,shorten left=5pt,kpoint common,fill=black]

\tikzstyle{kpointdag}=[kpoint adjoint]
\tikzstyle{kpointadj}=[kpoint adjoint]
\tikzstyle{kpointconj}=[kpoint conjugate]
\tikzstyle{kpointtrans}=[kpoint transpose]

\tikzstyle{big kpoint}=[kpoint, minimum width=1.2 cm, minimum height=8mm, inner sep=4pt, text depth=3mm]

\tikzstyle{wide kpoint}=[kpoint, minimum width=1 cm, inner sep=2pt, text depth=-0.7 mm]
\tikzstyle{wide kpointdag}=[kpointdag, minimum width=1 cm, inner sep=2pt, text depth=0.7 mm]
\tikzstyle{wide kpointconj}=[kpointconj, minimum width=1 cm, inner sep=2pt, text depth=-0.7 mm]
\tikzstyle{wide kpointtrans}=[kpointtrans, minimum width=1 cm, inner sep=2pt, text depth=0.7 mm]

\tikzstyle{gray kpoint}=[kpoint,fill=gray!50!white]
\tikzstyle{gray kpointdag}=[kpointdag,fill=gray!50!white]
\tikzstyle{gray kpointadj}=[kpointadj,fill=gray!50!white]
\tikzstyle{gray kpointconj}=[kpointconj,fill=gray!50!white]
\tikzstyle{gray kpointtrans}=[kpointtrans,fill=gray!50!white]

\tikzstyle{gray dkpoint}=[kpoint,fill=gray!50!white,doubled]
\tikzstyle{gray dkpointdag}=[kpointdag,fill=gray!50!white,doubled]
\tikzstyle{gray dkpointadj}=[kpointadj,fill=gray!50!white,doubled]
\tikzstyle{gray dkpointconj}=[kpointconj,fill=gray!50!white,doubled]
\tikzstyle{gray dkpointtrans}=[kpointtrans,fill=gray!50!white,doubled]

\tikzstyle{white label}=[draw,fill=white,rectangle,inner sep=0.7 mm]
\tikzstyle{gray label}=[draw,fill=gray!50!white,rectangle,inner sep=0.7 mm]
\tikzstyle{black label}=[draw,fill=black,rectangle,inner sep=0.7 mm]

\tikzstyle{dkpoint}=[kpoint,doubled]
\tikzstyle{wide dkpoint}=[wide kpoint,doubled]
\tikzstyle{dkpointdag}=[kpoint adjoint,doubled]
\tikzstyle{dkcopoint}=[kpoint adjoint,doubled]
\tikzstyle{dkpointadj}=[kpoint adjoint,doubled]
\tikzstyle{dkpointconj}=[kpoint conjugate,doubled]
\tikzstyle{dkpointtrans}=[kpoint transpose,doubled]

\tikzstyle{kscalar}=[kpoint common, shape=EBox, inner xsep=-1pt, inner ysep=3pt,font=\small]
\tikzstyle{kscalarconj}=[kpoint common, shape=WBox, inner xsep=-1pt, inner ysep=3pt,font=\small]

 \tikzstyle{upground}=[circuit ee IEC,thick,ground,rotate=90,scale=2.5]
 \tikzstyle{downground}=[circuit ee IEC,thick,ground,rotate=-90,scale=2.5]
 \tikzstyle{bigground}=[regular polygon,regular polygon sides=3,draw=gray,scale=0.50,inner sep=-0.5pt,minimum width=10mm,fill=gray]

\tikzstyle{arrs}=[-latex,font=\small,auto]
\tikzstyle{arrow plain}=[arrs]
\tikzstyle{arrow dashed}=[dashed,arrs]
\tikzstyle{arrow bold}=[very thick,arrs]
\tikzstyle{arrow hide}=[draw=white!0,-]
\tikzstyle{arrow reverse}=[latex-]
\tikzstyle{cdnode}=[]

\usepackage{tikz-qtree}
\usepackage{tikz-qtree-compat}

\title{A Corpus-based Toy Model for DisCoCat}
\author{
	Stefano Gogioso
	\institute{Quantum Group \\ University of Oxford}
	\email{stefano.gogioso@cs.ox.ac.uk}
}
\def\titlerunning{A Corpus-based Toy Model for DisCoCat}
\def\authorrunning{S. Gogioso}

\begin{document}

\maketitle

\begin{abstract}
The categorical compositional distributional (DisCoCat) model of meaning rigorously connects distributional semantics and pregroup grammars, and has found a variety of applications in computational linguistics. From a more abstract standpoint, the DisCoCat paradigm predicates the construction of a mapping from syntax to categorical semantics. In this work we present a concrete construction of one such mapping, from a toy model of syntax for corpora annotated with constituent structure trees, to categorical semantics taking place in a category of free $R$-semimodules over an involutive commutative semiring $R$.
\end{abstract}

\section{Introduction} 

The paradigm of distributional semantics draws its roots in the \textit{distributional hypothesis} of \cite{harris1968mathematical}, evocatively captured by the following words of John Rupert Firth \cite{firth1957papers}:
\begin{center}
\textit{You shall know a word by the company it keeps.}
\end{center}
In modern computational linguistics, distributional semantics is characterised by the application of statistical methods to large corpora of data. While suitable for certain classes of words (such as nouns), a vanilla statistical approach fails to capture the compositional aspects of language, such as those involving words acting as modifiers (e.g. adjectives and determiners), or words mediating interactions (e.g. verbs and prepositions). 

The compositional aspect alone is well covered by the pregroup approach to syntax and grammar of Lambek \cite{lambek1958mathematics,lambek1999type}. The categorical compositional distributional (DisCoCat) model of meaning \cite{coecke2010mathematical} connects distributional semantics and pregroup grammars, exploiting a close connection between the latter and compact-closed symmetric monoidal categories. The categorical formalism allows for the transfer of tools and diagrammatics from categorical quantum mechanics \cite{abramsky2009categorical}.

Frobenius algebras are key structures in categorical quantum mechanics, where they define orthonormal bases and observables. Through the connection established by the DisCoCat framework, they have also found a variety of applications in computational linguistics. They have been used to model, amongst other things, adjectives and verbs \cite{kartsaklis2013reasoning}, relative pronouns \cite{sadrzadeh2013frobenius,sadrzadeh2014frobenius} and intonational / informational structure \cite{kartsaklis2015frobenius}. More generally, they provide an efficient way to represent and manipulate operators on an $M$-dimensional vector space in time/space linear in $M$ (rather than quadratic). The extension of categorical quantum mechanics from the pure-state case to the mixed-state case via the CPM construction \cite{selinger2007dagger} has also found application in linguistics: in recent work, density matrices have been used to model ambiguity \cite{piedeleu2015open} and sentence entailment \cite{balkir2016distributional}.

In this work we construct an abstract categorical model for DisCoCat starting from a generic corpus\footnote{From our abstract standpoint, semantics depend entirely one the given corpus, no matter the size or quality of annotations. Larger, better annotated corpora will result in better semantics, smaller or poorly annotated corpora will result in worse semantics.} annotated with constituent structure trees. Concretely, we will work with context-free grammars \`{a} la Chomsky \cite{chomsky1995minimalist}, but Combinatory Categorial Grammar (CCG) \cite{steedman2000syntactic} and dependency grammars \cite{kubler2009dependency} could also be used. In Section \ref{section_syntax} we construct a toy model of syntax for the corpus, as a simplifying intermediate step between constituent structure trees and the categorical semantics. In our toy model of syntax, words are classified according to three possible functions: 
\begin{enumerate}
	\item[(i)] \textit{object words}: the basic building blocks (modelled as vectors in the semantic space);
	\item[(ii)] \textit{modifier words}: modify individual object words (modelled as unary operators on the space); 
	\item[(iii)] \textit{interaction words}: connect distinct object words (modelled as binary operators on the space).
\end{enumerate}
In Section \ref{section_semantics}, we consider the compact-closed symmetric monoidal category $\RModCategory{R}$ of $R$-semimodules over some involutive commutative semiring $R$. We model object words as vectors in a free $R$-semimodule $\SpaceH$, which we construct from our intermediate toy model of syntax. Using Frobenius algebras, we model modifier words as unary operators on $\SpaceH$, and interaction words as binary operators on $\SpaceH$. In Section \ref{section_conclusions}, we discuss some possible future extensions of this model. 

\newcommand{\corpus}{\mathcal{C}}
\newcommand{\len}[1]{\operatorname{len}\, #1}
\newcommand{\CatLex}{\operatorname{Cat}_{\operatorname{L}}}
\newcommand{\CatPhr}{\operatorname{Cat}_{\operatorname{P}}}
\newcommand{\sentence}{{\underline{s}}}
\newcommand{\sentencew}[1]{{s_{#1}}}
\newcommand{\wordw}{{\overline{w}}}
\newcommand{\word}{\wordw}
\newcommand{\wordu}{{\overline{u}}}
\newcommand{\wordv}{{\overline{v}}}
\newcommand{\treeroot}[1]{root\left[#1\right]}
\newcommand{\objectSet}[1]{\mathcal{O}\left[#1\right]}
\newcommand{\instancesIn}[1]{\widehat{W}\left[#1\right]}
\newcommand{\wordsIn}[1]{W\left[#1\right]}
\newcommand{\completion}[1]{\overline{#1}}
\newcommand{\modifiers}[1]{M\left[#1\right]}
\newcommand{\interactions}[2]{I\left[#1,#2\right]}
\newcommand{\domainInfluence}[1]{D\left[#1\right]}
\newcommand{\objectWords}{{ObjW}}

\section{The toy model of syntax}
\label{section_syntax}

We begin by considering an abstract annotated corpus. We model this as a set $\corpus$ of independent sentences, each sentence $\sentence$ coming with an annotated tree $T(\sentence)$ describing its grammatical structure. Concretely, we will have in mind the constituent structure trees from context-free grammars \`{a} la Chomsky, as in the following example:
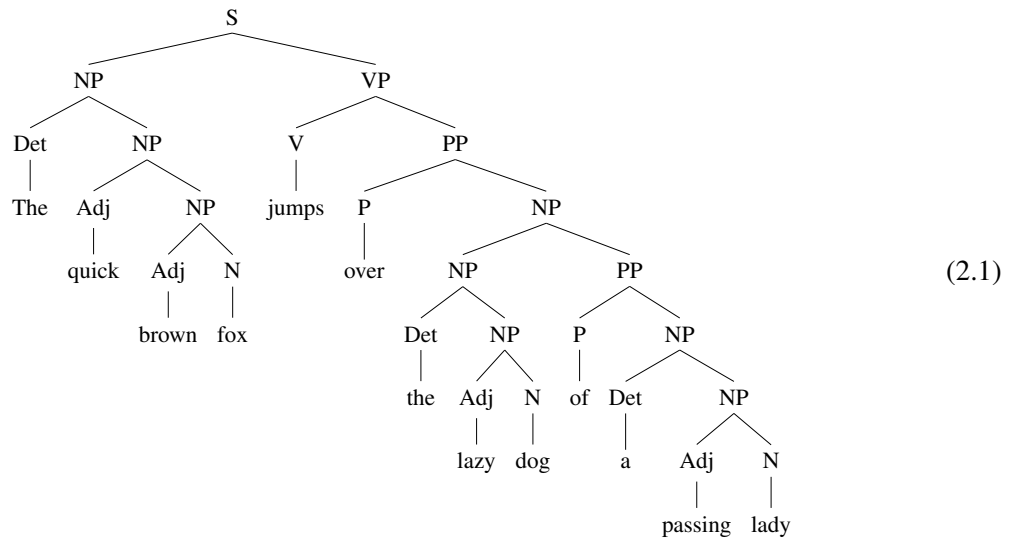
\begin{equation}\label{exampleTree}
\begin{multlined}
\begin{tikzpicture}[scale = 1.6]
\tikzset{frontier/.style={distance from root=240pt}}
\Tree 
[.S
	[.NP
		[.Det
			\text{The}
		]
		[.NP
			[.Adj
				\text{quick}
			]
			[.NP
				[.Adj
					\text{brown}
				]
				[.N
					\text{fox}
				]
			]
		]
	]
	[.VP
		[.V
			\text{jumps}
		]
		[.PP
			[.P
				\text{over}
			]
			[.NP
				[.NP
					[.Det
						\text{the}
					]
					[.NP
						[.Adj
							\text{lazy}
						]
						[.N
							\text{dog}
						]
					]
				]
				[.PP
					[.P
						\text{of}
					]
					[.NP
						[.Det
							\text{a}
						]
						[.NP
							[.Adj
								\text{passing}
							]
							[.N
								\text{lady}
							]
						]
					]
				]
			]
		]
	]
]
\end{tikzpicture}
\end{multlined}
\end{equation}
We assume the internal nodes to be labelled from a finite set of \textbf{phrasal categories}, and the leaf nodes (corresponding to the words of the sentence) to be labelled from a finite set of \textbf{lexical categories}; we will refer to the set of all possible labels, both for internal nodes and for leaf nodes, as the set of \textbf{syntactic categories}. Concretely, we will have the following categories in mind:
\begin{enumerate}
\item[-] \textbf{Lexical categories: } Adjectives (Adj), Prepositions (P), Adverbs (Adv), Conjunctions (Conj), Determiners (Det), Nouns (N), Personal Pronouns (Pron), Possessive Pronouns (Poss), Verbs (V).
\item[-] \textbf{Phrasal categories: } Adjective Phrases (AdjP), Adverb Phrases (AdvP), Adposition Phrases (PP), Noun Phrases (NP), Verb Phrases (VP), Sentences (S).
\end{enumerate}
This set of categories has been chosen for simplicity of exposition: realistic applications would use a richer set (e.g. Penn Treebank tagset). With minor modifications, constituency grammar could be substituted with Combinatory Categorial Grammar (CCG)\footnote{Additional functional annotations might be required to adapt this model to CCG.}; similarly, dependency grammars could be used.

We assume the constituent structure trees to be binary, with the exception of nodes involving conjunctions (which we assume to be ternary with sub-trees on the left / right and a leaf of lexical category Conj in the middle). A proper treatment of conjunctions (both coordinating and subordinating) does not appear here, and is left to future work. Leaf nodes will be labelled from the set $\CatLex$ of lexical categories, and internal nodes will be labelled from the set $\CatPhr$ of phrasal categories. If we denote by $W$ the set of all words appearing in the corpus, each sentence $\sentence$ (say $N_\sentence$ words long) will be modelled by a function $\sentence : \{1,...,N_\sentence\} \rightarrow W \times \CatLex$; because the corpus is annotated, there is no ambiguity in lexical categories for the words in a given sentence. When talking about words, we will henceforth mean pairs $\word := (w,c) \in W \times \CatLex$ of a (bare) word tagged with a lexical category. We will denote individual instances of words in sentences of the corpus by $(\sentence,j)$, meaning the $j$-th word instance of sentence $\sentence \in \corpus$.

\noindent We will distinguish between three possible functions\footnote{The functions below are \textit{defined} in the remainder of this section. They are not necessarily related to other works.} for words in a sentence:
\begin{enumerate}
\itemsep0em
\item[(i)] the \textbf{object words}, the ones which will be modelled as vectors in the semantic space;
\item[(ii)] the \textbf{modifier words}, which alter the meaning of individual object words;
\item[(iii)] the \textbf{interaction words}, which connect the meaning of distinct object words.
\end{enumerate}
We fix a set of \textbf{object categories}, both lexical and phrasal, and we declare the \textbf{object words} to be those with lexical category in this set (let the set of object words be denoted by $\objectWords$). Concretely, we will take the object categories to be the lexical categories N and Pron\footnote{This is not entirely satisfactory. Treatment of personal pronouns in our framework will certainly require additional efforts.}, and the phrasal categories NP and S. 

The reason which induced us to include the phrasal category S of sentences into the list of object categories goes as follows. In this work, everything will revolve around a semantic space $\SpaceH$ for objects, with no semantic space for sentences. From the point of view of the model presented here, the noun phrase \textit{a jumping fox} and the sentence \textit{a fox jumps} will be treated equally\footnote{Assuming that \textit{jumping} and \textit{jumps} would be identified by lemmatisation of the corpus.}, in the sense that both will be taken to be statements about some generic fox which jumps. While the elimination of sentences might seem rather radical, it is consistent with the belief in a world made entirely of interacting objects.

Following this view, we define the \textbf{objects} of a sentence $\sentence$ (denote their set by $\objectSet{\sentence}$) to be the subtrees $T \leq T(\sentence)$ having an object category as root node; in particular, object words (the leaves of $T(\sentence)$, seen as singleton subtrees) will be objects of the sentence. We also denote by $\instancesIn{T}$ the set of word instances in $\sentence$ spanned by an object $T \leq T(\sentence)$, and by $\wordsIn{T}$ the corresponding set of words $\suchthat{\word \in W \times \CatLex}{\word = \sentencew{j} \text{ and } (\sentence,j) \in \instancesIn{T}}$. The set of objects $\objectSet{\sentence}$ is a partially ordered set $(\objectSet{\sentence},\leq)$ under the subtree ordering $\leq$ inherited from $T(\sentence)$; in fact, it has the structure of a tree (as long as we assume that the root of $T(\sentence)$ is always of object category). In the following figure, the poset of objects for sentence (\ref{exampleTree}) is shown. Some objects are given labels (numerals in square brackets) for ease of representation.
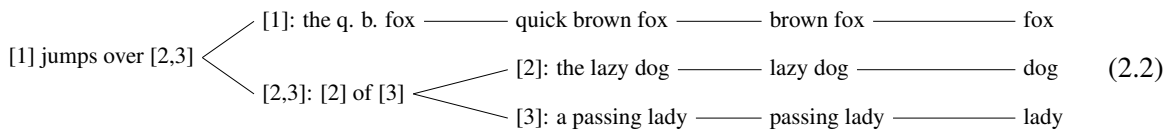
\begin{equation}\label{objectPoset}
\begin{multlined}
\begin{tikzpicture}[scale = 1.6]
\tikzset{grow'=right}
\tikzset{execute at begin node=\strut}
\tikzset{every tree node/.style={anchor=base west}}
\tikzset{level distance=120pt}
\Tree
[.\text{[1] jumps over [2,3]}
	[.\text{[1]: the q. b. fox}
		[.\text{quick brown fox}
			[.\text{brown fox}
				\text{fox}
			]
		]
	]
	[.\text{[2,3]: [2] of [3]}
		[.\text{[2]: the lazy dog}
			[.\text{lazy dog}
				\text{dog}
			]
		]
		[.\text{[3]: a passing lady}
			[.\text{passing lady}
				\text{lady}
			]
		]
	]
]
\end{tikzpicture}
\end{multlined}
\end{equation}

\noindent Given an object $T \in \objectSet{\sentence}$, we define its \textbf{completion} $\completion{T}$ to be the largest $T' \geq T$ such that every $T'' \leq T'$ intersects $T$. In the example (\ref{objectPoset}) above, we have $\completion{dog}$ = \textit{the lazy dog}, $\completion{lady}$ = \textit{a passing lady}, and $\completion{fox}$ = \textit{the quick brown fox}. The completion of an object $T$ in a sentence is the largest object in the sentence which contains $T$ and does not contain any object disjoint from $T$: it completes $T$ to its most detailed description in $\sentence$ not involving interactions with independent objects. 

We have seen that the objects of sentence are organised in a hierarchy of increasing detail: we wish a modifier word (instance) for an object $T$ to be one which applies to $T$ alone, making it more specific in the sentence. Following this intuition, we define the set $\modifiers{T}$ of \textbf{modifiers}/\textbf{modifier words} of $T$ to be 
\begin{equation}
\modifiers{\,T\,} := \instancesIn{\,\completion{T}\,} - \instancesIn{\,T\,}
\end{equation}

\noindent In the example (\ref{objectPoset}) above, the modifiers of \textit{brown fox} are $\{the,quick\}$, specifying that the object \textit{brown fox} in the sentence is a specific one, and quick as well; similarly, the modifiers of $lady$ are $\{a,passing\}$, specifying that the object $lady$ in the sentence is a generic one, and is passing by. It should be mentioned that this approach is intrinsically intersective in nature: this will reflect in a modelling of modifier words as commuting unary operators\footnote{We observe in passing that non-intersective, contextual behaviour in linguistics requires, at the level of semantics, the same operational features required by contextuality in quantum theory, namely non-commutativity of operators.}.

We characterised modifier words for an object $T$ to be those word (instances) which apply to it alone, making it more specific. Dually, we wish to characterise interaction words for two objects $T,T'$ to be those words which are \underline{needed} to put $T$ and $T'$ in relation in the sentence; in particular, modifiers of $T$ and $T'$ should not appear as interactions. Given two objects $T,T' \in \objectSet{\sentence}$, we write $T \vee T'$ to denote the \textbf{join} (or most recent common ancestor) of $T$ and $T'$ in the tree $\objectSet{\sentence}$. In the example (\ref{objectPoset}) above, $dog \vee lady$ = \textit{the lazy dog of a passing lady}, while both $fox \vee lady$ and $fox \vee dog$ are the entire sentence (and so is $fox \vee dog \vee lady$). We shall henceforth require that every object in every sentence of the corpus spans at least an object word: under this assumption, it is always true that a sentence is the completion of the join of all its object words. Following the intuition given above, we define the set $\interactions{T}{T'}$ of \textbf{interactions}/\textbf{interaction words} between $T$ and $T'$ to be 
\begin{equation}
\interactions{\,T}{T'\,} := \instancesIn{\,T\vee T'\,} - \instancesIn{\,\completion{T}\,} - \instancesIn{\,\completion{T'}\,}
\end{equation}

\noindent It should be noted that, in accordance with the intuition given above for completions, we have that $\interactions{\,T}{T'\,} = \interactions{\,\completion{T}}{\completion{T'}\,}$ (because $T \vee T' = \completion{T} \vee \completion{T'}$). In the example (\ref{objectPoset}) above, the interactions of $dog$ and $lady$ form the singleton set $\{of\}$, indicating that the only relationship required to connect $dog$ and $lady$ is one of possession. Moreover, the interactions of $fox$ and $dog \vee lady$ form the set $\{jumps,over\}$, indicating that \textit{the quick brown fox} and \textit{the lazy dog of a passing lady} are put in relation by an action of jumping and some vertical distance\footnote{One of the main disadvantages of the intersective approach used here is the issue of phrasal verbs in the English language. A more sophisticated approach, with a more sophisticated grammar, might declare unbreakable phrasal units to cope with this issue (and with some more general instances of non-intersective modifiers).}. In order to force compatibility with the left-right ordering of words in pregroup grammars and DisCoCat, we will henceforth set $\interactions{\,T}{T'\,} := \emptyset$ unless $T$ appears on the left of $T'$ (as disjoint subtrees of $T(\sentence)$).

We conclude this section by recapping the fundamentals of our toy model for syntax. We have fixed a set of \textit{object categories}, and defined \textit{object words} as words (word instances, to be more precise) from those categories. Based entirely on the respective constituent structure tree, we have given a definition of \textit{objects} in a sentence (subsuming object words are a special case), of \textit{modifier words} (extending objects to more detailed objects) and of \textit{interaction words} (connecting distinct objects).

\section{The categorical compositional distributional model}
\label{section_semantics}

In this section we turn the toy model of semantics defined in the previous section\footnote{Consisting of object words, modifier words, and interaction words, together with their hierarchical structure.} into a categorical compositional distributional model. We fix an involutive\footnote{I.e. a commutative semiring with a chosen involution $^\star$, with similar axioms to conjugation in the semiring/field $\complexs$.} commutative semiring $(R,^\star)$, and consider the category $\RModCategory{R}$ of finite-dimensional free $R$-semimodules. The objects of this category are all in the form $R^X$ for some finite set $X$, and morphisms $R^X \rightarrow R^Y$ can be represented as matrices in $R^{Y \times X}$, exactly as in the real and complex cases (with $R=\reals$ and $R=\complexs$ respectively). The category $\RModCategory{R}$ is a dagger symmetric monoidal category, with dagger and tensor product defined exactly as in the real and complex cases. Similarly, it is a dagger compact category: as a consequence, it is a suitable model category for categorical compositional distributional semantics. For sake of simplicity, we will restrict ourselves to the case $R = \naturals$, with the trivial involution $n^\star := n$. A more sophisticated choice of semiring and involution would give us additional freedom in the modelling of words as vectors and $R$-linear operators: additional semiring elements could be used to signal polarity, modality and/or inflection of word instances; we leave this to future work. It should also be noted that the semantic model presented here is a free one, with a basis ranging over all instances of words in the corpus. Future work will see the development of categorically-sensible compression techniques, both for the purposes of real-world implementation and to enable the emergence of richer semantics from the restriction of available degrees of freedom.

\subsection{The distributional part}

Consider the set $X := \bigcup_{\sentence \in \corpus} \{\sentence\} \times \suchthat{j \in \{1,...,N_{\sentence}\}}{\sentencew{j} \in \objectWords}$ of all instances of object words in sentences of the corpus, and define the \textbf{semantic space for objects $\SpaceH$} to be the finite-dimensional free $R$-semimodule $R^X$. The $R$-semimodule $\SpaceH$ comes with a \textbf{standard orthonormal basis} $\big(\ket{\sentence,j}\big)_{(\sentence,j) \in X}$ indexed by object word instances. To each object word $\word \in \objectWords$ we associate a \textbf{semantic vector} $\ket{\word}$ in $\SpaceH$, defined as follows on the standard orthonormal basis:
\begin{equation} 
\begin{tikzpicture}[node distance = 1cm]
\node[point] (w) {$\word$};
\node (out) [above of = w, yshift = -0.25cm] {};
\node (equal) [right of = w, yshift = 0.2cm]{$:=$};
\node (def) [right of = equal, xshift = -0.1cm, yshift = -0.1cm] {$\mathlarger{\sum}\limits_{\sentencew{j} = \word} $};
\node[point,inner sep = -0mm] (v) [right of = def, xshift = 0.3cm, yshift = -0.1cm]{$\sentence,j$};
\node (outv) [above of = v, yshift = -0.25cm] {};
\begin{pgfonlayer}{background}
\draw[-] (w) to (out);
\draw[-] (v) to (outv);
\end{pgfonlayer}
\end{tikzpicture}
\end{equation}

\newcommand{\cooccurInnerProd}[1]{IP_{#1}}
\noindent An object word $\word$ is then modelled as the indicator function of its instances in the corpus. In particular, different words are assigned orthogonal vectors, and the squared norm $\braket{\word}{\word} \in R$ of a word counts the total number of occurrences of $\word$ throughout the corpus. In other approaches, the inner product of the vectors associated with two words is used to encode some notion of semantic distance between them. One such notion, that of co-occurrence (within a window of $2k+1$ object words, with $k$ fixed), can be easily recovered in the framework presented here (and so can many other linear notions of distance).

Recall that the inner product can be obtained as $\braket{\word}{\word} = \cap_{\SpaceH} \cdot \big(\ket{\word}^\star \otimes \ket{\word}\big)$, where the bilinear map $\cap: \SpaceH \otimes \SpaceH \rightarrow R$ is the usual cap\footnote{The dagger compact category $\RModCategory{R}$ has self-dual objects, where the conjugate $\ket{\psi}^\star \in R^Y$ of a vector $\ket{\psi} \in R^Y$ is obtained by application of the involution $^\star$ to each coordinate of $\ket{\psi}$ in the standard orthonormal basis of $R^Y$.} given by the dagger compact structure. The familiar co-occurrence picture can be recovered by considering a different bilinear map $\cooccurInnerProd{k}: \SpaceH \otimes \SpaceH \rightarrow R$, which counts the co-occurrences $\cooccurInnerProd{k} \cdot \big( \ket{\word}^\star \otimes \ket{\word'}\big) \in R$ of two object words $\word,\word'$ in windows of $2k+1$ object words within the same sentence:
\begin{equation} 
\begin{tikzpicture}[node distance = 1cm]
\node[box, minimum width = 1.5cm] (IP) {$\cooccurInnerProd{k}$};
\node (inl) [below of = IP, yshift = +0.25cm, xshift = -0.5cm] {};
\node (inr) [below of = IP, yshift = +0.25cm, xshift = +0.5cm] {};
\node (inla) [above of = inl] {};
\node (inra) [above of = inr] {};
\node (equal) [right of = IP, xshift = 0.4cm, yshift = 0.1cm]{$:=$};
\node (def) [right of = equal, xshift = 0.2cm, yshift = -0.2cm] {$\mathlarger{\sum}\limits_{\substack{(\sentence,j),(\sentence,j') \\ \text{s.t. } 0 \leq |i-j| \leq k}}$};
\node[copoint,inner sep = -0.22mm] (v) [right of = def, xshift = 1cm, yshift = +0.1cm]{$\sentence,j$};
\node (outv) [below of = v, yshift = +0.25cm] {};
\node[copoint,inner sep = -0.6mm] (v2) [right of = v, xshift = 0.2cm]{$\sentence,j'$};
\node (outv2) [below of = v2, yshift = +0.25cm] {};
\begin{pgfonlayer}{background}
\draw[-] (v) to (outv);
\draw[-] (v2) to (outv2);
\draw[-] (inl) to (inla);
\draw[-] (inr) to (inra);
\end{pgfonlayer}
\end{tikzpicture}
\end{equation}

\newcommand{\modifier}[1]{M_{#1}}
\newcommand{\interaction}[1]{I_{#1}}
\subsection{The compositional part}

Now that object words have been associated to vectors in the semantic space $\SpaceH$, we will model modifiers and interactions as operators. When $\SpaceH$ is very high-dimensional (as is certainly the case here), operators $\SpaceH \rightarrow \SpaceH$ can be extremely expensive to concretely work with: as a consequence, one often restricts the attention to a more convenient subclass of operators, obtained by using Frobenius algebras. No matter what semiring $R$ we choose, the standard orthonormal basis of $\SpaceH$ always gives rise to a special commutative $\dagger$-Frobenius algebra $(\hbox{\begin{tikzpicture} [scale=0.8,transform shape] 

\def\deltax{0.3} 
\def\deltay{0.5} 


\node (mult_label_inl) at (-\deltax,-\deltay) {};
\node (mult_label_inr) at (+\deltax,-\deltay) {};
\node [dot, fill=\Zbwcolour] (mult) at (0,0) {};
\node (mult_label_out) at (0,+\deltay) {};

\draw[-] [out=90,in=225](mult_label_inl) to (mult);
\draw[-] [out=90,in=315](mult_label_inr) to (mult);
\draw[-] (mult) to (mult_label_out);

\end{tikzpicture}
}\!,\hbox{\begin{tikzpicture} [scale=0.8,transform shape] 

\def\deltax{0.3} 
\def\deltay{0.5} 

\path[use as bounding box] (-\deltax,-\deltay) rectangle (\deltax,\deltay);

\node [dot, fill=\Zbwcolour] (mult) at (0,-0.25*\deltay) {};
\node (mult_label_out) at (0,+\deltay) {};
\draw[-] (mult) to (mult_label_out);

\end{tikzpicture}
}\!,\hbox{\begin{tikzpicture} [scale=0.8,transform shape] 

\def\deltax{0.3} 
\def\deltay{0.5} 


\node (mult_label_outl) at (-\deltax,+\deltay) {};
\node (mult_label_outr) at (+\deltax,+\deltay) {};
\node [dot, fill=\Zbwcolour] (mult) at (0,0) {};
\node (mult_label_in) at (0,-\deltay) {};
\draw[-] [in=270,out=135] (mult) to (mult_label_outl);
\draw[-] [in=270,out=45] (mult) to (mult_label_outr);
\draw[-] (mult_label_in) to (mult);

\end{tikzpicture}
}\!,\hbox{\begin{tikzpicture} [scale=0.8,transform shape] 

\def\deltax{0.3} 
\def\deltay{0.5} 

\path[use as bounding box] (-\deltax,-\deltay) rectangle (\deltax,\deltay);

\node [dot, fill=\Zbwcolour] (mult) at (0,0.25*\deltay) {};
\node (mult_label_in) at (0,-\deltay) {};
\draw[-] (mult_label_in) to (mult);

\end{tikzpicture}
}\!)$ in the following way:
\begin{equation} 
\begin{tikzpicture}[node distance = 0.75cm]
\node[dot,fill=\Zbwcolour] (comult) {};
\node (in) [above of = comult] {};
\node (outl) [below of = comult, xshift = -0.5cm] {};
\node (outr) [below of = comult, xshift = 0.5cm] {};
\node (eq) [right of = comult] {$:=$};
\node (sum) [right of = eq, yshift = -0.1cm] {$\mathlarger{\sum}\limits_{(\sentence,j)\in X}$};
\node[copoint,inner sep = -0.38mm] (v) [right of = sum, xshift = 0.5cm, yshift = +0.1cm]{$\sentence,j$};
\node (outv) [below of = v, yshift = -0.0cm] {};
\node[copoint,inner sep = -0.38mm] (v2) [right of = v, xshift = 0.7cm]{$\sentence,j$};
\node (outv2) [below of = v2, yshift = -0.0cm] {};
\node[point,inner sep = -0.38mm] (v1) [right of = v, xshift = -0.0cm, yshift = 0.2cm]{$\sentence,j$};
\node (outv1) [above of = v1, yshift = -0.0cm] {};
\begin{pgfonlayer}{background}
\draw[-] (in) to (comult);
\draw[-,out=225,in=90] (comult) to (outl);
\draw[-,out=315,in=90] (comult) to (outr);
\draw[-] (v) to (outv);
\draw[-] (v1) to (outv1);
\draw[-] (v2) to (outv2);
\end{pgfonlayer}
\node[dot,fill=\Zbwcolour] (comult) [right of = comult, xshift = 6cm] {};
\node (in) [above of = comult] {};
\node (outl) [below of = comult, xshift = -0.5cm] {};
\node (outr) [below of = comult, xshift = 0.5cm] {};
\node (eq) [right of = comult] {$:=$};
\node (sum) [right of = eq, yshift = -0.1cm] {$\mathlarger{\sum}\limits_{(\sentence,j)\in X}$};
\node[point,inner sep = -0.38mm] (v1) [right of = sum, xshift = 0.5cm, yshift = 0.2cm]{$\sentence,j$};
\node (outv1) [above of = v1, yshift = -0.0cm] {};
\begin{pgfonlayer}{background}
\draw[-] (in) to (comult);
\draw[-] (v1) to (outv1);
\end{pgfonlayer}
\end{tikzpicture}
\end{equation}
 
\noindent One then considers the family of operators $\SpaceH \rightarrow \SpaceH$ taking the following form, which can be efficiently\footnote{If $\SpaceH$ is $M$-dimensional, then these operators admit an $M$-dimensional representation, compared with the $M^2$-dimensional representation required in general by operators $\SpaceH \rightarrow \SpaceH$.} represented by means of vectors in $\SpaceH$:
\begin{equation} \label{diagonalisedOperator}
\begin{tikzpicture}[node distance = 1cm]
\node[dot,fill=\Zbwcolour] (dot) {};
\node (in) [below of = dot] {};
\node (out) [above of = dot, yshift = -0.25cm] {};
\node[point] (state) [below right of = dot] {$\psi$};
\begin{pgfonlayer}{background}
\draw[-] (in) to (out);
\draw[-,out=90,in=315] (state) to (dot);
\end{pgfonlayer}
\node (eq) [right of = dot, xshift = 0.0cm] {$=$};
\node (def) [right of = eq, xshift = 1.1cm] {$\ket{\sentence,j} \;\;\mapsto\;\; \psi_{(\sentence,j)}\ket{\sentence,j}$};
\node (eq) [left of = dot, xshift = -0cm]{$P_\psi\;\;:=$};
\end{tikzpicture}
\end{equation}
Operators in this form can be easily manipulated using the algebra operations: for example, if $\ket{\psi \odot \phi} := \hbox{\begin{tikzpicture} [scale=0.8,transform shape] 

\def\deltax{0.3} 
\def\deltay{0.5} 


\node (mult_label_inl) at (-\deltax,-\deltay) {};
\node (mult_label_inr) at (+\deltax,-\deltay) {};
\node [dot, fill=\Zbwcolour] (mult) at (0,0) {};
\node (mult_label_out) at (0,+\deltay) {};

\draw[-] [out=90,in=225](mult_label_inl) to (mult);
\draw[-] [out=90,in=315](mult_label_inr) to (mult);
\draw[-] (mult) to (mult_label_out);

\end{tikzpicture}
}\! \cdot (\ket{\psi} \otimes \ket{\phi})$ denotes the algebra multiplication, then composition of operators in the form above is given by $P_{\phi} \circ P_{\psi} = P_{\psi \odot \phi}$ (using associativity of the algebra multiplication).

A particularly interesting class of operators $\SpaceH \rightarrow \SpaceH$ is given by the \textbf{projectors}, the self-adjoint idempotent operators. It is easy to show\footnote{Observing that $(\psi \odot \phi)_{(\sentence,j)} = \psi_{(\sentence,j)} \phi_{(\sentence,j)}$ in $R$.} that an operator in the form (\ref{diagonalisedOperator}) above is a projector if and only if all the coordinates $\psi_{(\sentence,j)}$ are self-conjugate idempotents in the semiring $R$. In the case of fields (such as $R=\reals$ or $R=\complexs$), or semirings with cancellation (such as $R=\integers$ or $R=\naturals$), the only idempotents in $R$ are $0$ and $1$; however, more general semirings will have more (self-conjugate) idempotents, giving us more projectors.

Modifiers were defined to be words which alter the meaning of individual object words by making it more specific: following this intuition, we model them as projectors $\SpaceH \rightarrow \SpaceH$. Given a non-object word $\wordu$, we define the associated \textbf{modifier} $\modifier{\wordu}: \SpaceH \rightarrow \SpaceH$ to be the projector over the subspace given by all the objects which $\wordu$ modifies in the corpus:
\begin{equation} \label{modifierProjector}
\begin{tikzpicture}[node distance = 1cm]
\node[dot,fill=\Zbwcolour] (dot) {};
\node (in) [below of = dot] {};
\node (out) [above of = dot, yshift = -0.25cm] {};
\node[point] (state) [below right of = dot] {$m_\wordu$};
\begin{pgfonlayer}{background}
\draw[-] (in) to (out);
\draw[-,out=90,in=315] (state) to (dot);
\end{pgfonlayer}
\node (eq) [right of = dot, xshift = 1cm] {where};
\node[point] (w) [right of = eq, yshift = -0.5cm, xshift = 1cm] {$m_\wordu$};
\node (out) [above of = w, yshift = -0.25cm] {};
\node (equal) [right of = w, yshift = 0.2cm]{$:=$};
\node (def) [right of = equal, xshift = -0.1cm, yshift = -0.1cm] {$\mathlarger{\sum}\limits_{(\sentence,j) \in m_{\wordu}} $};
\node[point,inner sep = -0mm] (v) [right of = def, xshift = 0.3cm, yshift = -0.1cm]{$\sentence,j$};
\node (outv) [above of = v, yshift = -0.25cm] {};
\node (eq) [left of = dot, xshift = -0cm]{$\modifier{\wordu}\;\;:=$};
\begin{pgfonlayer}{background}
\draw[-] (w) to (out);
\draw[-] (v) to (outv);
\end{pgfonlayer}
\end{tikzpicture}
\end{equation}
and we defined $m_{\wordu} := \suchthat{(\sentence,j) \in X}{\exists T \in \objectSet{\sentence} \text{ with } \wordu \in \modifiers{T} \text{ and } (\sentence,j) \leq T}$ to be the set of instances of object words which appear in objects modified by $\wordu$. A modifier $\modifier{\wordu}$ sends the semantic vector $\ket{\word}$ of an object word $\word$ to the indicator function $\modifier{\wordu} \ket{\word}$ of all the instances of $\word$ appearing in objects modified by $\wordu$.

\noindent In their formulation as projectors, modifiers come with some interesting logical structure (in the language of quantum theory, the structure of a commutative algebra of projectors). If the semiring $R$ satisfies the additive cancellation law (i.e. if $a=b$ whenever $a+c = b+c$ for some $c \in R$), then it is possible to define a partial subtraction operation on elements by setting $b-a := c$  whenever a $c$ exists such that $b = a+c$ (if it exists, it is necessarily unique). In $\naturals$, this returns the usual subtraction (and similarly in $\integers$ with full-fledged additive inverses). When $R$ satisfies the additive cancellation law, modifiers come with natural logical operations:
\begin{enumerate}
\item[(i)] if $\wordu$ and $\wordv$ are two non-object words, then $\modifier{\wordu} \circ \modifier{\wordv}$ (which always equals $\modifier{\wordv} \circ \modifier{\wordu}$ since $(\hbox{\begin{tikzpicture} [scale=0.8,transform shape] 

\def\deltax{0.3} 
\def\deltay{0.5} 


\node (mult_label_inl) at (-\deltax,-\deltay) {};
\node (mult_label_inr) at (+\deltax,-\deltay) {};
\node [dot, fill=\Zbwcolour] (mult) at (0,0) {};
\node (mult_label_out) at (0,+\deltay) {};

\draw[-] [out=90,in=225](mult_label_inl) to (mult);
\draw[-] [out=90,in=315](mult_label_inr) to (mult);
\draw[-] (mult) to (mult_label_out);

\end{tikzpicture}
}\!,\hbox{\begin{tikzpicture} [scale=0.8,transform shape] 

\def\deltax{0.3} 
\def\deltay{0.5} 

\path[use as bounding box] (-\deltax,-\deltay) rectangle (\deltax,\deltay);

\node [dot, fill=\Zbwcolour] (mult) at (0,-0.25*\deltay) {};
\node (mult_label_out) at (0,+\deltay) {};
\draw[-] (mult) to (mult_label_out);

\end{tikzpicture}
}\!,\hbox{\begin{tikzpicture} [scale=0.8,transform shape] 

\def\deltax{0.3} 
\def\deltay{0.5} 


\node (mult_label_outl) at (-\deltax,+\deltay) {};
\node (mult_label_outr) at (+\deltax,+\deltay) {};
\node [dot, fill=\Zbwcolour] (mult) at (0,0) {};
\node (mult_label_in) at (0,-\deltay) {};
\draw[-] [in=270,out=135] (mult) to (mult_label_outl);
\draw[-] [in=270,out=45] (mult) to (mult_label_outr);
\draw[-] (mult_label_in) to (mult);

\end{tikzpicture}
}\!,\hbox{\begin{tikzpicture} [scale=0.8,transform shape] 

\def\deltax{0.3} 
\def\deltay{0.5} 

\path[use as bounding box] (-\deltax,-\deltay) rectangle (\deltax,\deltay);

\node [dot, fill=\Zbwcolour] (mult) at (0,0.25*\deltay) {};
\node (mult_label_in) at (0,-\deltay) {};
\draw[-] (mult_label_in) to (mult);

\end{tikzpicture}
}\!)$ is commutative) can equivalently be obtained as $P_{m_{\wordu} \odot m_{\wordv}}$, and is a well-defined projector\footnote{Frobenius algebras thus play a very active role in information flow within the sentence, combining modifier words together into larger modifiers. Later on, they will also be seen to combine interaction words together into larger interactions.}. It projects onto the subspace given by all object word instances appearing in objects modified by \underline{both} $\wordu$ and $\wordv$. We will also write this as $\modifier{\wordu} \odot \modifier{\wordv}$. 
\item[(ii)] if $\wordu$ and $\wordv$ are two non-object words, then $\modifier{\wordu} + \modifier{\wordv} - \modifier{\wordu} \odot \modifier{\wordv}$ is always a well defined projector. It projects on the subspace given by all object word instances appearing in objects modified by \underline{at least one} of $\wordu$ and $\wordv$. We will write this as $\modifier{\wordu} \oplus \modifier{\wordv}$.
\item[(iii)] if $\wordu$ is a non-object word, then $\id{\SpaceH}-\modifier{\wordu}$ is always a well defined projector. It projects onto the subspace given by all object word instances appearing in objects \underline{not} modified by $\wordu$. We will write this as $1\ominus \modifier{\wordu}$. 
\end{enumerate}

\noindent Interactions were defined to be words which connect the meaning of different object words within the same sentence: we model them as operators $\SpaceH \otimes \SpaceH \rightarrow \SpaceH$. Given a word $\wordu$ not of object category, we define the associated \textbf{interaction} $\interaction{\wordu}: \SpaceH \otimes \SpaceH \rightarrow \SpaceH$ to be the following binary operation:
\begin{equation} \label{interactionOperator}
\begin{tikzpicture}[node distance = 1cm]
\node[dot,fill=\Zbwcolour] (dot) {};
\node (in) [below of = dot] {};
\node (out) [above of = dot, yshift = -0.25cm] {};
\node[point] (state) [below right of = dot, yshift = -0.25cm] {$i_\wordu$};
\begin{pgfonlayer}{background}
\draw[-] (in) to (out);
\draw[-,out=90,in=315] (state.60) to (dot);
\end{pgfonlayer}
\node (eq) [right of = dot, xshift = 1cm] {where};
\node[point] (w) [right of = eq, yshift = -0.5cm, xshift = 1cm] {$i_\wordu$};
\node (outl) [above of = w, yshift = -0.25cm, xshift = -0.1cm] {};
\node (outr) [above of = w, yshift = -0.25cm, xshift = +0.1cm] {};
\node (equal) [right of = w, yshift = 0.2cm]{$:=$};
\node (def) [right of = equal, xshift = -0.1cm, yshift = -0.1cm] {$\mathlarger{\sum}\limits_{\big((\sentence,j),(\sentence,j')\big) \in i_{\wordu}} $};
\node[point,inner sep = -0mm] (v) [right of = def, xshift = 0.8cm, yshift = -0.1cm]{$\sentence,j$};
\node (outv) [above of = v, yshift = -0.25cm, xshift = 0.3cm] {};
\node[point,inner sep = -0.45mm] (v2) [right of = v, xshift = 0.2cm]{$\sentence,j'$};
\node (outv2) [above of = v2, yshift = -0.25cm, xshift = -0.3cm] {};
\begin{pgfonlayer}{background}
\draw[-,out=90,in=270] (w.120) to (outl);
\draw[-,out=90,in=270] (w.60) to (outr);
\draw[-,out=90,in=270] (v) to (outv);
\draw[-,out=90,in=270] (v2) to (outv2);
\end{pgfonlayer}
\node[dot,fill=\Zbwcolour] (dot) [left of = dot, xshift = 0.5cm] {};
\node (in) [below of = dot] {};
\node (out) [above of = dot, yshift = -0.25cm] {};
\begin{pgfonlayer}{background}
\draw[-] (in) to (out);
\draw[-,out=90,in=315] (state.120) to (dot);
\end{pgfonlayer}
\node[box, minimum width = 1.2cm] (plus) [right of = out, xshift = -0.75cm] {$+$};
\node (out) [above of = plus, yshift = -0.25cm] {};
\begin{pgfonlayer}{background}
\draw[-] (plus) to (out);
\end{pgfonlayer}
\node (eq) [left of = dot, xshift = -0cm]{$\interaction{\wordu}\;\;:=$};
\end{tikzpicture}
\end{equation}
and we defined $i_{\wordu} := \suchthat{\big((\sentence,j),(\sentence,j')\big) \in X \times X}{\exists T,T' \in \objectSet{\sentence} \text{ with } \wordu \in \interactions{T}{T'} \text{ and } (\sentence,j) \leq T, (\sentence,j') \leq T'}$ to be the set of pairs of instances which appear in objects put into relation by $\wordu$. The interaction $\interaction{\wordu}$ is obtained as a projector on $\SpaceH \otimes \SpaceH$, selecting pairs of instances in $i_{\wordu}$, followed by the linear operator\\\frame{$\;\mathlarger{+}\;$} $:\SpaceH \otimes \SpaceH \rightarrow \SpaceH$ defined as follows:
\begin{equation}
\text{\frame{$\;\mathlarger{+}\;$}} \; := \Big(\id{\SpaceH} \otimes \hbox{\begin{tikzpicture} [scale=0.8,transform shape] 

\def\deltax{0.3} 
\def\deltay{0.5} 

\path[use as bounding box] (-\deltax,-\deltay) rectangle (\deltax,\deltay);

\node [dot, fill=\Zbwcolour] (mult) at (0,0.25*\deltay) {};
\node (mult_label_in) at (0,-\deltay) {};
\draw[-] (mult_label_in) to (mult);

\end{tikzpicture}
}\! + \hbox{\begin{tikzpicture} [scale=0.8,transform shape] 

\def\deltax{0.3} 
\def\deltay{0.5} 

\path[use as bounding box] (-\deltax,-\deltay) rectangle (\deltax,\deltay);

\node [dot, fill=\Zbwcolour] (mult) at (0,0.25*\deltay) {};
\node (mult_label_in) at (0,-\deltay) {};
\draw[-] (mult_label_in) to (mult);

\end{tikzpicture}
}\! \otimes \id{\SpaceH}\Big) = \ket{\sentence,j} \otimes \ket{\sentence',j'} \mapsto \ket{\sentence,j} + \ket{\sentence',j'} 
\end{equation}
The interaction $\interaction{\wordu}$ sends a pair $\word,\word'$ of object words to the vector $\ket{\psi}$ spanning all object words appearing in objects containing instances of $\word$ and $\word'$ and related by $\wordu$:
\begin{equation}\label{interactionActingOnWords}
\interaction{\wordu} \cdot \Big( \ket{\word} \otimes \ket{\word'} \Big)\;\; = \sum_{\big((\sentence,j),(\sentence,j')\big) \in i_{\wordu}(\word,\word')} \Big(\ket{\sentence,j}+\ket{\sentence,j'} \Big)
\end{equation}
where we defined $i_{\wordu}(\word,\word') := \suchthat{\big((\sentence,j),(\sentence,j')\big) \in i_{\wordu}}{\word = \sentencew{j}\text{ and }\word' = \sentencew{j}}$. Note that the instances $(\sentence,j)$ and $(\sentence,j')$ appearing in the sum of Equation (\ref{interactionActingOnWords}) are necessarily distinct even when $\word = \word'$.

\noindent Contrary to modifiers, interactions associated with two non-object words $\wordu,\wordv$ cannot be composed directly to obtain an interaction, because of the $\text{\frame{$\;\mathlarger{+}\;$}}$ map compressing $\SpaceH \otimes \SpaceH$ to $\SpaceH$. However, the projector parts of the interactions do compose, and the joint interaction $\interaction{\wordu} \odot \interaction{\wordv}$ can be defined by using the Frobenius multiplication $\hbox{\begin{tikzpicture} [scale=0.8,transform shape] 

\def\deltax{0.3} 
\def\deltay{0.5} 


\node (mult_label_inl) at (-\deltax,-\deltay) {};
\node (mult_label_inr) at (+\deltax,-\deltay) {};
\node [dot, fill=\Zbwcolour] (mult) at (0,0) {};
\node (mult_label_out) at (0,+\deltay) {};

\draw[-] [out=90,in=225](mult_label_inl) to (mult);
\draw[-] [out=90,in=315](mult_label_inr) to (mult);
\draw[-] (mult) to (mult_label_out);

\end{tikzpicture}
}\! \otimes \hbox{\begin{tikzpicture} [scale=0.8,transform shape] 

\def\deltax{0.3} 
\def\deltay{0.5} 


\node (mult_label_inl) at (-\deltax,-\deltay) {};
\node (mult_label_inr) at (+\deltax,-\deltay) {};
\node [dot, fill=\Zbwcolour] (mult) at (0,0) {};
\node (mult_label_out) at (0,+\deltay) {};

\draw[-] [out=90,in=225](mult_label_inl) to (mult);
\draw[-] [out=90,in=315](mult_label_inr) to (mult);
\draw[-] (mult) to (mult_label_out);

\end{tikzpicture}
}\!$ on $\SpaceH \otimes \SpaceH$:
\begin{equation} \label{MultipleInteractionOperator}
\begin{tikzpicture}[node distance = 1cm]
\node[dot,fill=\Zbwcolour] (dot) {};
\node (in) [below of = dot] {};
\node (out) [above of = dot, yshift = -0.25cm] {};
\node[point] (statel) [below right of = dot, yshift = -0.75cm, xshift = 0.0cm] {$i_\wordu$};
\node[point] (stater) [below right of = dot, yshift = -0.75cm, xshift = 1.0cm] {$i_\wordv$};
\node[dot,fill=\Zbwcolour] (dotl) [above of = statel, yshift = -0.5cm] {};
\node[dot,fill=\Zbwcolour] (dotr) [above of = stater, yshift = -0.5cm] {};
\begin{pgfonlayer}{background}
\draw[-] (in) to (out);
\draw[-,out=90,in=0] (dotr) to (dot);
\draw[-,out=90,in=225] (statel.120) to (dotl.225);
\draw[-,out=90,in=180] (statel.60) to (dotr.180);
\draw[-,out=90,in=0] (stater.120) to (dotl.0);
\draw[-,out=90,in=315] (stater.60) to (dotr.315);
\end{pgfonlayer}
\node[dot,fill=\Zbwcolour] (dot) [left of = dot, xshift = 0.5cm] {};
\node (in) [below of = dot] {};
\node (out) [above of = dot, yshift = -0.25cm] {};
\begin{pgfonlayer}{background}
\draw[-] (in) to (out);
\draw[-,out=90,in=315] (dotl) to (dot);
\end{pgfonlayer}
\node[box, minimum width = 1.2cm] (plus) [right of = out, xshift = -0.75cm] {$+$};
\node (out) [above of = plus, yshift = -0.25cm] {};
\begin{pgfonlayer}{background}
\draw[-] (plus) to (out);
\end{pgfonlayer}
\node (eq) [left of = dot, xshift = -0.25cm]{$\interaction{\wordu} \odot \interaction{\wordv}\;\;:=$};
\end{tikzpicture}
\end{equation}

\noindent Finally, the following figure presents the full semantic vector associated with the original sentence from example (\ref{exampleTree}). Associativity and commutativity of the Frobenius algebra multiplication has been used to group modifiers and interactions together, improving readability.
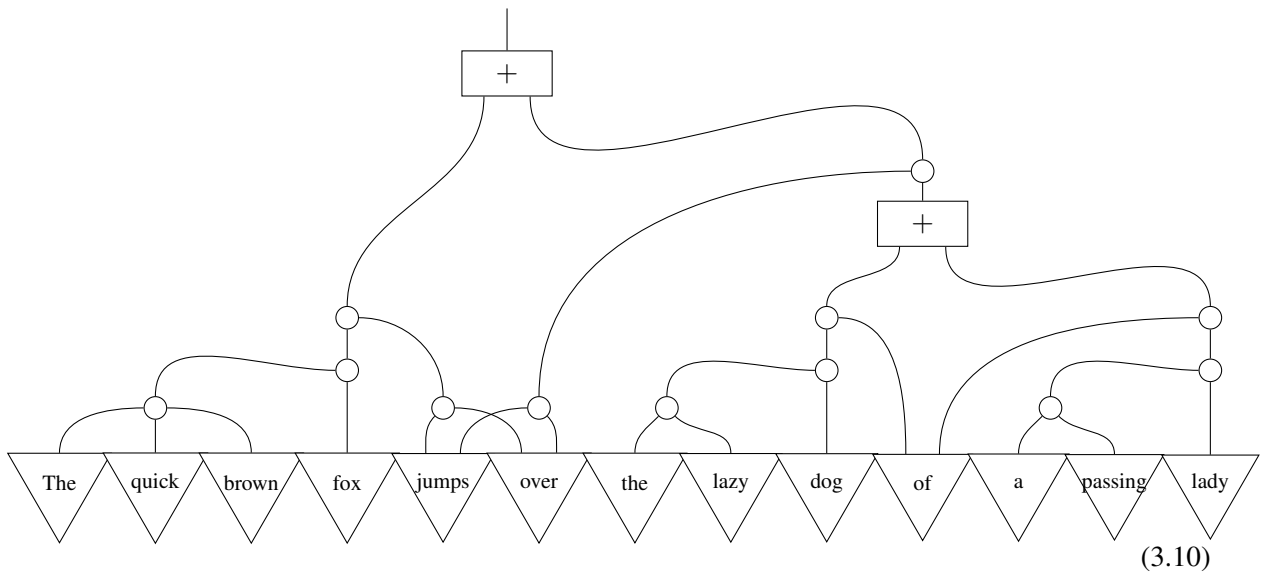
\begin{equation} 
\begin{tikzpicture}[node distance = 1.7cm]
\node[point,inner sep = 0.75mm] (The) {The};
\node[point,inner sep = -0.5mm] (quick) [right of = The] {quick};
\node[point,inner sep = -1.1mm] (brown) [right of = quick] {brown};
\node[point,inner sep = 1.2mm] (fox) [right of = brown] {fox};
\node[point,inner sep = -1mm] (jumps) [right of = fox] {jumps};
\node[point,inner sep = 0.4mm] (over) [right of = jumps] {over};
\node[point,inner sep = 1.4mm] (the) [right of = over] {the};
\node[point,inner sep = 0.5mm] (lazy) [right of = the] {lazy};
\node[point,inner sep = 0.8mm] (dog) [right of = lazy] {dog};
\node[point,inner sep = 2mm] (of) [right of = dog] {of};
\node[point,inner sep = 2.75mm] (a) [right of = of] {a};
\node[point,inner sep = -2.2mm] (passing) [right of = a] {passing};
\node[point,inner sep = 0.2mm] (lady) [right of = passing] {lady};
\node[dot,fill=\Zbwcolour] (Tqb) [above of = quick, yshift = -0.7cm] {};
\node[dot,fill=\Zbwcolour] (Tqbf) [above of = fox, yshift = -0.2cm] {};
\node[dot,fill=\Zbwcolour] (tl) [above of = lazy, yshift = -0.7cm, xshift = -0.85cm] {};
\node[dot,fill=\Zbwcolour] (tld) [above of = dog, yshift = -0.2cm] {};
\node[dot,fill=\Zbwcolour] (ap) [above of = passing, yshift = -0.7cm, xshift = -0.85cm] {};
\node[dot,fill=\Zbwcolour] (apl) [above of = lady, yshift = -0.2cm] {};
\begin{pgfonlayer}{background}
\draw[-,out=90,in=180] (The) to (Tqb);
\draw[-,out=90,in=270] (quick) to (Tqb);
\draw[-,out=90,in=0] (brown) to (Tqb);
\draw[-,out=90,in=225] (the) to (tl);
\draw[-,out=90,in=315] (lazy) to (tl);
\draw[-,out=90,in=225] (a) to (ap);
\draw[-,out=90,in=315] (passing) to (ap);
\draw[-,out=90,in=270] (fox) to (Tqbf);
\draw[-,out=90,in=180] (Tqb) to (Tqbf);
\draw[-,out=90,in=270] (dog) to (tld);
\draw[-,out=90,in=180] (tl) to (tld);
\draw[-,out=90,in=270] (lady) to (apl);
\draw[-,out=90,in=180] (ap) to (apl);
\end{pgfonlayer}
\node[dot,fill=\Zbwcolour] (tldo) [above of = tld, yshift = -1cm] {};
\node[dot,fill=\Zbwcolour] (oapl) [above of = apl, yshift = -1cm] {};
\node[box, minimum width = 1.2cm] (tldoapl) [above of = of, yshift = 1.75cm] {$+$};
\begin{pgfonlayer}{background}
\draw[-,out=90,in=270] (tld) to (tldo);
\draw[-,out=90,in=0] (of.120) to (tldo);
\draw[-,out=90,in=270] (apl) to (oapl);
\draw[-,out=90,in=180] (of.60) to (oapl);
\draw[-,out=90,in=270] (tldo) to (tldoapl.225);
\draw[-,out=90,in=270] (oapl) to (tldoapl.315);
\end{pgfonlayer}
\node[dot,fill=\Zbwcolour] (j) [above of = jumps, yshift = -0.7cm] {};
\node[dot,fill=\Zbwcolour] (o) [above of = over, yshift = -0.7cm] {};
\begin{pgfonlayer}{background}
\draw[-,out=90,in=225] (jumps.120) to (j);
\draw[-,out=90,in=180] (jumps.60) to (o);
\draw[-,out=90,in=0] (over.120) to (j);
\draw[-,out=90,in=315] (over.60) to (o);
\end{pgfonlayer}
\node[dot,fill=\Zbwcolour] (Tqbfjo) [above of = Tqbf, yshift = -1cm] {};
\node[dot,fill=\Zbwcolour] (jotldoapl) [above of = tldoapl, yshift = -1cm] {};
\node[box, minimum width = 1.2cm] (Tqbfjotldoapl) [above of = jumps, yshift = 3.75cm, xshift = 0.85cm] {$+$};
\node (out) [above of = Tqbfjotldoapl, yshift = -0.7cm] {};
\begin{pgfonlayer}{background}
\draw[-,out=90,in=270] (tldoapl) to (jotldoapl);
\draw[-,out=90,in=180] (o) to (jotldoapl);
\draw[-,out=90,in=270] (Tqbf) to (Tqbfjo);
\draw[-,out=90,in=0] (j) to (Tqbfjo);
\draw[-,out=90,in=270] (Tqbfjo) to (Tqbfjotldoapl.225);
\draw[-,out=90,in=270] (jotldoapl) to (Tqbfjotldoapl.315);
\draw[-] (Tqbfjotldoapl) to (out);
\end{pgfonlayer}
\end{tikzpicture}
\end{equation}

\section{Future work}
\label{section_conclusions}

We look forward to improve and extend the model in the following directions. First of all, a more sophisticated choice of commutative semiring $R$ would give us additional semantic degrees of freedom: more idempotents could be used to identify the different roles played by words in interactions (e.g. by introducing a \textit{subject} idempotent and an \textit{object} idempotent), while a self-inverse element could be used to distinguish positive occurrences of object words from negated ones. Second, the category $\RModCategory{R}$ admits a CPM construction, which can be used to model ambiguous semantics. Finally, the construction given here is a \textit{free} one, with basis ranging over all instances of words in the corpus: for efficient real-world applications, a suitable compressed construction should be devised. Also, the case of conjunctions needs to be fully treated, and a better modelling of personal pronouns is deemed necessary.
\vspace{-0.25cm}

\subparagraph*{Acknowledgements} 
The author would like to thank Bob Coecke for suggestions, comments and useful discussions, as well as Sukrita Chatterji and Nicol\`o Chiappori for their continued support. Funding from EPSRC and Trinity College is gratefully acknowledged. 

\bibliographystyle{eptcs}
\bibliography{biblio}

\end{document}